# THE SINGULARITY CONTROVERSY

Part I:

## Lessons Learned and Open Questions:
## Conclusions from the Battle on the Legitimacy of the Debate

Amnon H. Eden

Sapience Project 2016

### SYNOPSIS

This report seeks to inform policy makers on the nature and the merit of the arguments for and against the concerns associated with a potential technological singularity.

Part I describes the lessons learned from our investigation of the subject, separating the arguments of merit from the fallacies and misconceptions that confuse the debate and undermine its rational resolution.

ACM Computing Classification System (CCS 2012): Computing methodologies~Philosophical/theoretical foundations of artificial intelligence • Social and professional topics~Codes of ethics

# CONTENTS

London, United Kingdom

Sapience Project is a thinktank dedicated to the study of disruptive and intelligent computing. Its charter is to identify, extrapolate, and anticipate disruptive, long-lasting and possibly unintended consequences of progressively intelligent computation on economy and society; and to syndicate focus reports and mitigation strategies.

## BOARD

Vic Callaghan, University of Essex

B. Jack Copeland, University of Canterbury

Amnon H. Eden, Sapience Project

Jim Moor, Dartmouth College

David Pearce, BLTC Research

Steve Phelps, Kings College London

Anders Sandberg, Future of Humanity Institute, Oxford University

Tony Willson, Helmsman Services

# MOTIVATION

Lately artificial intelligence (AI) has been receiving unusual attention. News outlets scream "Celebrated Scientists: The End is Nigh!" and tell of prospects of a "robot uprising" catastrophe that artificial intelligence may bring about. Others dismiss these concerns as speculative, cultish, apocalyptic "Rapture of the Nerds" nonsense. An American think-tank has gone as far as calling Elon Musk and Stephen Hawking (and, by implication, some of computer science's foremost geniuses historically!) "fear-mongering luddites", "innovation killers" and a danger to technological progress and to America's safety.

It is not unusual for scientific controversies to occupy headlines. What's unusual about the criticism of AI concerns is that it delegitimises the debate. The attacks on expressions of concern over AI risks, or over how it should be researched and used, have create the (largely inaccurate) impression that experts disagree whether the subject has significance. Such strong and conflicting opinions by authority figures are confusing.

We ask: Who is right? Should we fear "killer robots"? Stop research in AI? Or perhaps all concerns to public safety are just headline-grabbing, click-bait claptrap? Is the debate at all legitimate?

The prospects of a major disruptive event —a *technological singularity* — by around 2045 has been examined by scientists for decades and much further back by philosophers. Unfortunately, the recent debate has not always been informed and misunderstandings, misdirection, and misinformation have created the illusion of a bitter dispute.

The general public has also been exposed to personal assistants (eg Siri), satellite navigation, picture tagging, game 'bots' and other applications of AI. "Singularity" blockbusters and popular TV series brought new (and old) ideas to the general public. And a trend in popular media articles with Arnold Schwarzenegger's cyborg face in *Terminator* depicts a dumbed down, misinformed, and often mocked perceptions of a "singularity", which muddy the controversy further.

In reality, like the so called "debates" on climate change and genetic engineering, domain experts are much more in agreement concerning the timeframe of human-level AI. Where disagreements between AI experts exist, they concern scenarios far subtler and more complex than the "robot uprising" fairytale. But few can dedicate their time to understand the complex questions that arise from what may be wrongly considered purely technical questions. Fewer yet can keep up with the barrage of publications in the field.

Yet if there is any truth to these concerns, we cannot afford a "policy vacuum".

This is the first report in a series on the singularity controversy, written to inform policy and decisions makes on the lessons that Fellows of the Project Sapience have learned from studying a potential technological singularity since 2009, including the publication of the first peer-reviewed collection on superintelligence: the Singularity Hypotheses volume (Eden et. al 2013).

The first report focuses on false and unfounded arguments which have repeatedly been used to dismiss AI concerns and delegitimise the debate. We believe that the pernicious misconceptions which have haunted the debate undermine its rational resolution. Refuting these fallacies is therefore our first step.

# STUDYING THE SINGULARITY

Although thoughts about the singularity may appear to be very new, in fact such ideas have a long philosophical history. To help increase awareness of the deep roots of singularitarian thought within traditional philosophy, it may be useful to look at some of its historical antecedents.

Many philosophers portrayed cosmic process as an ascending curve of positivity. Over time, the quantities of intelligence, power or value are always increasing. Technological versions have sometimes invoked broad technical progress and have sometimes focused on more specific outcomes such as the possible recursive self-improvement of artificial intelligence. Historical analysis of a broad range of paradigm shifts in science, biology, history, and technology — in particular in computing technology — suggest an accelerating rate of progress.

The predictive power of biological evolution, cultural evolution, and technological evolution have been attempted to be unified the under the titles "Big History" (Christian 2010; Christian 2012) and the "Law of Accelerating Returns" (Kurzweil 2004). Consequently, John von Neumann forecasted the arrival of an "essential singularity in the history of the race beyond which human affairs as we know them could not continue": a statement taken to coin the term technological singularity in it's current use.

This notion of singularity coincides in time and nature with Alan Turing's (1950) expectation of machines to exhibit intelligence on a par with the average human by 2050. John Irving Good (Good 1965) and Vernor Vinge (1993) expect it to take the form of an 'intelligence explosion': the process by which ultra-intelligent machines design ever more intelligent machines.

Critics of the technological singularity dismiss these claims as speculative and empirically unsound, if not pseudo-scientific. Many valid arguments have been put forth, but by 2009 our enquiry has not produced satisfying results because it has remained unclear what the arguments under attack actually are. We have learned that accounts of a technological singularity appear to disagree on its causes and possible consequences, on timescale and even on the nature of superintelligence to emerge, namely machine or (post-) human? An event or a period? Is the technological singularity unique, or have there been others? The absence of a consensus on basic questions casts doubt whether the notion of *singularity* is at all coherent.

To clarify the issues we organised the conference track "Technological Singularity and Acceleration Studies" at the European Conference for Computing and Philosophy (ECAP), inviting researchers to present contributions from a philosophical, computational, mathematical, and scientific points of view (Eden 2009). In July 2009, researchers from the Singularity Institute (later renamed the Machine Learning Research Institute), Dartmouth College and elsewhere presented their work at the venue of the 7th ECAP at the Autonomous University of Barcelona, and again in October 2010 (Eden 2010) for the 8th ECAP at the Technical University in Munich. Discussions sought to promote the debate in singularity arguments, for example using methods for probabilistic risk assessment, and examining the plausibility of AI arms race scenarios by comparing it with the Cold War's scenarios. None of the track's participants argued that AI risks should be ignored. The main conclusion has been that the issues deserve further clarification, and that an appropriate debate on merit of singularity hypotheses had not taken place yet.

Therefore, in 2010 we called prominent experts with opposing views in the debate to address each other in a collection of essays (Eden et al. 2010) entitled "The Singularity Hypothesis: A Scientific and Philosophical Assessment". Virtually all scientists and philosopher who's studied the subject were invited to make a contribution to the book. By the beginning of 2012 the essays were developed and collected into a volume.

To promote the debate further we've invited a short rebuttal to each essay from known critics. The essays appeared in the volume published in 2013 (Eden, Moor, et al. 2013), each followed by a short rebuttal. The rebuttals, unique among volumes in similar topics, has proven to be a crucial contribution to the debate. Below we list some of our conclusions.

# CONCLUSIONS DRAWN

Almost a decade of research in these issues has led us to the following conclusions:

## Conclusion (A):
## Singularity = Acceleration + Discontinuity + Superintelligence

The term *technological singularity* in its contemporary sense traces back to John von Neumann (in Ulam, 1958), popularized by Vernor Vinge (Vinge 1993) and Ray Kurzweil (Kurzweil 2006), and further elaborated by Eliezer Yudkowsky (Yudkowsky 2007), Robin Henson (Hanson 2008), and David Chalmers (Chalmers 2010).

The differences in the accounts have raised doubt whether the term is coherent. Our investigation has led us to believe that despite their differences, all careful expositions of a technological singularity occurring by the mid-21st century can be uniquely described using three common characteristics: *superintelligence*[1], *acceleration*, and *discontinuity*.

**Superintelligence** underlies all singularity scenarios, which see the quantitative measure of *intelligence*, at least as it is measured by traditional IQ tests (such as Wechsler and Stanford-Binet), becoming meaningless for capturing the intellectual capabilities of minds that are orders of magnitude more intelligent than us. Alternatively, we may say a graph measuring average intelligence beyond the singularity in terms of IQ score may display some form of radical discontinuity if superintelligence emerges. This remains true whether a particular account describes the AI or the IA scenario (see Conclusion (A)). For example, it is commonly argued that the arrival of human-level AI may soon be followed by artificial superintelligence. And accounts of a singularity from human amplification describe superhuman cognitive capabilities, including unbounded memory and accelerating recall times, and the eradication of common obstructions to intelligent behaviour such as limited resources, disease and age.

The **discontinuity** in accounts of the singularity take the term to stand for a turning-point in human history, as in Von Neumann's canonical definition ("some essential singularity in the history of the race beyond which human affairs, as we know them, could not continue.") Whether it is taken to last a few hours (e.g. a 'hard takeoff' or 'FOOM' scenarios) or a period spread over decades (e.g., 'wave' (Toffler 1980)), the singularity term appears to originate not from mathematical singularities (Hirshfeld 2011) but from the discontinuities of *gravitational singularities* — better known as black holes. Seen as a central metaphor, a gravitational singularity is a (theoretical) point at the centre of black holes at which quantities that are otherwise meaningful (e.g., *density* and *space-time curvature*) become infinite, or rather meaningless. This metaphor is useful to express the idea that *superintelligence* stands for that level of intelligence in which traditional measures become ineffective of meaningless.

---

[1] The definition we gave in (Eden, Steinhart, et al. 2013) omits *superintelligence*, taking it to be implied from the exposition of singularity as the immergence of superintelligence, artificial or posthuman.

The *discontinuity* is best explained using the event horizon surrounding gravitational singularities: we cannot understand or foresee the events that may follow a singularity in the way we cannot peek into black holes before crossing the *event horizon*: a boundary in space-time (marked by the Schwarzschild radius around the gravitational singularity) beyond which events inside this area cannot be observed from outside, and a horizon beyond which gravitational pull becomes so strong that nothing can escape, even light (hence "black")—a point of no return. The emergence of *superintelligence* marks a similar point, an event horizon, because even "a tiny increment in problem-solving ability and group coordination is why we left the other apes in the dust" (Sandberg 2014): a discontinuity in our ontological and epistemological account of our existence.

**Acceleration** refers to a rate of growth in some quantity such as intelligence (Good 1965), computations per second per fixed dollar (Kurzweil, 2005), economic measures of *growth rate* (Hanson 1994; Bostrom 2014; Miller 2013) or total output of goods and services (Toffler, 1970), and *energy rate density* (Chaisson 2013). Other accounts of acceleration describe quantitative measures of physical, biological, social, cultural, and technological processes of evolution: milestones or paradigm shifts whose timing demonstrates an accelerating pace of change. For example, Carl Sagan's Cosmic Calendar (1977: ch. 1) names milestones in biological evolution such as the emergence of eukaryotes, vertebrates, amphibians, mammals, primates, hominidae, and *Homo sapiens*, which show an accelerating trend. Some authors attempt to unify the acceleration in these quantities under one law of nature (Adams 1904; Kurzweil 2004; Christian 2010; Chaisson 2013): quantitatively or qualitatively measured, acceleration is commonly visualized as an upwards-curved mathematical graph which, if projected into the future, is said to be leading to a *discontinuity* (see above).

## Conclusion (B):

### Which singularity do you mean? AI or IA?

We conclude above that all singularity accounts have three common (and seemingly unique) denominators. Nonetheless, it has also become abundantly clear that singularity hypotheses refer to either one of two distinct and very different scenarios:

One scenario, the **singularity of artificial (super-)intelligence** (AI), refers to (often dystopian) the emergence of 'human-level AI' and smarter, superintelligent agents or software-based synthetic minds within about four decades. Absent precautions, this scenario sees AI becoming one of the greatest potential threats to human existence (Sandberg 2014) or "the worst thing to happen to humanity in history" (Hawking et al. 2014). On the other hand, the **singularity of posthuman superintelligence**, however, refers to radically different (usually utopian) scenario which sees the emergence of superintelligent posthumans within a similar timeframe, resulting from the merger of our biology with technology (Kurzweil 2005), specifically intelligence amplification (IA). This singularity would see our descendants overcoming disease, aging, hunger, and most of the present sources of misery to humans.

In conclusion, the singularity is an ambiguous term, which either stands for one of the greatest threats to human existence (the AI scenario) or the diamaterically opposed humanity's most magnificent transcendence (the IA scenario). This ambiguity caused endless confusion which obscured the debate and helped undermining its legitimacy. The popular media gets the two singularities mixed up (Guardian and Cadwalladr 2014). Even respected magazine who undertaken to "debunk AI myths" (Ars Technica and Goodwins 2015) confuse the two scenarios.

The Singularity Hypotheses volume has been structured accordingly. In Part I computer scientists describe the uniquely 'intelligent' behaviour of programs that possess an ability to *learn*, the plausibility of an intelligence explosion, and the likelihood of a singularity of artificial superintelligence within four decades. In part II, computer scientists discussed the risks of artificial superintelligence and described the research in 'friendly AI'. Essays in Part III of the volume discussed human enhancement and the plausibility of a singularity of human superintelligence within similar timeframe.

## Conclusion (C) :
## Some 'singularities' are implausible, incoherent, or no singular

Part IV of the Singularity Hypotheses volume was dedicated to critics of the technological singularity. Many view it as a religious idea (Proudfoot 2013; Bringsjord, Bringsjord, and Bello 2013). The idea of artificial superintelligence emerging by about 2045 has been mockingly described as the 'rapture of the nerds' (Doctorow and Stross 2013): an (yet another) apocalyptic fantasy, a technology-infused variation of doom-and-gloom scenarios originating from mysticism, fiction, cults, and commercial interests.

A powerful counterargument to singularity hypotheses is offered in the *energy density rate* theory offered by the physicist Eric Chaisson (Chaisson 2013). Chaisson accepts *acceleration* and *superintelligence* — two of the three premises underlying singularity accounts (see Conclusion (A): Singularity = Acceleration + Discontinuity + Superintelligence), but rejects *discontinuity*. The most potent part of Chaisson's argument shows that one physical quantity (*energy density rate*) can define and demonstrate the evolution of physical (eg, galaxies), biological (eg, life), cultural and technological (eg, aeroplanes) systems, thereby unifying all forms of evolution. As such, Chaisson's work explains and coincides with that of Carl Sagan (Sagan 1977), the historians behind Big History (Christian 2010; Christian 2012; Nazaretyan 2013), the Law of Accelerating Returns (Kurzweil 2004) and others. Chaisson's theory also embraces the emergence of *superintelligence* as "the next evolutionary leap forward beyond sentient beings". But Chaisson's ambitious acceleration theory also underplays the premise of *discontinuity*: in cosmic scale, he argues, there is no reason to claim that this step be "any more important than the past emergence of increasingly intricate complex systems."

## Conclusion (D) :
## Pulp "singularities" are *not* scientific hypotheses

The term *singularity* in general use has been vastly influenced by the recent spate of Hollywood blockbusters and popular TV series. To illustrate their use of the term, philosophers (Schneider 2009) and scientists (Hawking et al. 2014) used movies to illustrate how they convey their understanding of *singularity*.

Indeed, some science fiction stories has greatly contributed to our understanding of AI behaviour and its risks. But blockbusters and TV series have also taken artistic freedom, drawing in-creasingly irrational scenarios. For example, the 'robot uprising' scenario in the film *Terminator*[2] depicts a malevolent artificial superintelligence ("Skynet") who sets killer robots to chase crowds and mercilessly gun them down. Similarly, AIs in *The Matrix* have enslaved the human race, farming our bodies[3] for the highly unlikely purpose of generating electricity[4].

The unprecedented popularity of cinematic works that used the term have associated the (technological) singularity with that variation of the *Golem* fairytale in which the human race struggles to survive against an army of hostile shapeshifting robots. Thus the cultural construct — the singularity *meme* (Dawkins 2000; Blackmore 2001) — has become synonymous with the pulp fiction's malevolent AIs in the robot uprising narrative, largely ignoring subtler and more rational AIs such as depicted in *2001: Space Odyssey* and *Ex Machina*.

Many expositions of *singularity hypotheses* discuss the risks of artificial superintelligence. Most essayists explicitly reject the robot uprising scenario. Nonetheless, critics often attack the plot of films, conflating the singularity *meme* with scientific hypotheses. Confusion has remained despite numerous headlines in the mass media 'debunking' common misconceptions. Some singularity critics have even gone as far as to misattribute the plot of films to authors who have taken to distance themselves from the singularity meme[5]. This confusion has often derailed the debate. Below we have undertaken to debunk several fallacies arising from this confusion.

In discussing the risks of AI, authors of scientifically or philosophically sound singularity hypotheses distance themselves further from the singularity meme by emphasising the dangers of *indifferent* rather than aggressively hostile AI. The distinction between malevolent and indifferent separates the meme from the scientific hypotheses. It is explained further in our next Conclusion.

## Conclusion (E) :
### The risks of AI arise from indifference, *not* malevolence

Nearly all *singularity hypotheses* — that is, the scientifically and philosophically informed expositions of the singularity — have taken pains to reject the robot uprising scenario. Aggressive malevolence is uniformly dismissed as apocalyptic fantasy, one which may have been (mistakenly?) called "science fiction". The idea of a spontaneous emergence of a class of hostile AIs is considered anthropomorphic, irrational, and hence highly implausible.

Some AI theorists argued that self-preservation may be a basic drive (Omohundro 2008)[6], or a emergent property of artificial superintelligence rather than being explicitly programmed feature (Bostrom 2014). But rational singularity accounts reject the idea of AIs that have somehow spontaneously had human traits 'popping into existence'. Rather, singularity hypotheses argue that the risks associated with artificial superintelligence concern *indifference*, not malevolence.

Indeed, the *anthropomorphic bias* a recurring theme in the literature on AI safety. Human bias, they explain, leads us to expect just AI agent to undergo a sudden and unexplained emergence of emotions, such as malevolence; capabilities, such as empathy; and needs, such as autonomy and self-expression. In fact, an existential risk far more serious comes from an advanced race of artificially superintelligent agents that *lack* anthropomorphic emotions and needs. Therefore, by Occam's razor if not strictly in terms of probability theory, we can safely ignore the assumption that they will develop human traits spontaneously.

---

[2] and *Chronicles of Sarah Connor* TV series

[3] And in the process creating a very detailed simulation of reality to keep us under

[4] The human body generates only as much electricity as a potato — but people cost a lot more to 'farm' than potatoes.

[5] Eg (Guardian and Cadwalladr 2014) and (Ars Technica and Goodwins 2015)

[6] Omohundro's argument however makes several assumptions. The premises include a restricted class of artificial superintelligence (self-improving, utility-driven etc.) under specific circumstances (eg under free market forces etc.)

In fact, there is an even greater risk that the emerging race of superintelligence may be lethally indifferent to the human race, regardless whether the superintelligence is artificial, posthuman, alien, or whatnot. To see why one need only witness how we, *Homo Sapiens*, have been indifferent to the welfare of less intelligent species. We farm pigs in cramped cages because we like bacon and not because we "seek to enslave" the species. And Sea Minks have gone extinct not because anyone "sought to exterminate" them but because they were hunted down. The wider is the intelligence gap, the more indifferent superintelligent beings may be towards humans. They may not hesitate to wipe out a city of ordinary humans more than we hesitate to destroy an anthill while building a new motorway.

Concerns for AI safety are not limited to the emergence of a race of artificially superintelligent agents for AIs. The canonical example of the *Paperclip Factory* illustrates how dangerous can superintelligence be, even if it is built by well-meaning programmers, unless well-meaning, 'friendly' or 'philanthropic' supergoals are built into it:

> *The risks in developing superintelligence include the risk of failure to give it the supergoal of philanthropy. One way in which this could happen is ... that a well-meaning team of programmers make a big mistake in designing its goal system. This could result, to return to the earlier example, in a superintelligence whose top goal is the manufacturing of paperclips, with the consequence that it starts transforming first all of earth and then increasing portions of space into paperclip manufacturing facilities. (Bostrom 2003)*

More subtly, *indifference* may also refer to the risk which exists even from well-meaning artificially developed superintelligence: that indifference which arises from ignorance of humanity's *core values*. A well-meaning artificial superintelligence may be dangerous, unless it is 'friendly' to humans (Yudkowsky 2013; Bostrom 2014), in particular towards core values such as freedom, autonomy, and self-expression. Otherwise, the risk is that an AI may conclude that the best thing for the human race would be to put us all under hallucinogenic drugs, or bring about some 'false utopia', namely "a state of affairs that we might [temporarily] judge as desirable but which in fact ...things essential to human flourishing have been irreversibly lost." (Bostrom 2003)

# Conclusion (F) :
## The risk of AI is essentially like the risk of *any* powerful technology

Which advice are offered by authors discussing risks from AI? The Future of Life Institute has recently published an open letter "from AI & Robotics Researchers" urging AI research to be subject to the same risk considerations as research in the physical sciences:

> *Just as most chemists and biologists have no interest in building chemical or biological weapons, most AI researchers have no interest in building AI weapons (Future of Life Institute 2015)*

This argument for AI safety is neither novel nor unusual. For example, atomic energy can be very dangerous. To prevent its misuse, international treaties limit the production of military-grade nuclear energy and the sale of machinery and radioactive material capable of producing atomic weapons. Warning against an AI arms race (Future of Life Institute 2015), the Future of Life Institute's open letter takes the comparison to nuclear energy to its limits:

> *Unlike nuclear weapons, [autonomous killer robots] require no costly or hard-to-obtain raw materials, so they will become ubiquitous and cheap for all significant military powers to mass-produced. It will only be a matter of time until they appear on the black market and in the hands of terrorists, dictators wishing to better control their populace, warlords wishing to perpetrate ethnic cleansing, etc.*

Another analogy is offered to the risks of certain experiments in genetic engineering, in particular involving recombinant DNA which may create deadly pathogens in a process that is much less costly to than producing atomic energy, as well as far more available and easier to misused, but with results no-less catastrophic. To stem the misuse of such technology, the Asilomar conference called for a voluntary moratorium on certain experiments, but it also offered safety guidelines and recommended that the research continues. The measures proved effective and the moratorium has been universally observed. A recent study found several analogies between the concerns expressed in the Asilomar conference to the concerns from AI (Grace 2015), such as the focus on relatively novel risks from a new technology and the complexity of reasoning about the risks[7].

The science fiction of Isaac Asimov — who was a scientist as much as he was an author — took for granted the requirement for safeguards on machine intelligence. Asimov explained that safeguards which "apply, as a matter of course, to every tool that human beings use" (Asimov 1983). Elsewhere Asimov compares his Laws to circuit interrupters:

> *A tool must not be unsafe to use. Hammers have handles and screwdrivers have hilts to help increase grip. ... [To] perform its function efficiently [a tool mustn't] harm the user. This is the entire reason ground-fault circuit interrupters exist. Any running tool will have its power cut if a circuit senses that some current is not returning to the neutral wire, and hence might be flowing through the user. (Asimov 1993, in Wikipedia: "Robot Visions")*

In conclusion, technology has always been a double-edged sword, and even AI becomes an existential threat, it won't be the first one. *The Sorcerer's Apprentice* — if fantasy is to help us get the point — offers a more compelling scenario than that of the raging terminators. The tale of the dangerous outcome of the apprentice's careless experiments demonstrates the simple fact that when potent forces fall into his inadequately trained and irresponsible hands, control over forces may easily be lost, to deadly effects. Or as Nick Bostrom has put it:

*We need to be careful about what we wish for from a superintelligence, because we might get it.* (Bostrom 2003)

## Conclusion (G) :

## We're not clear what "artificial intelligence" means

Textbooks typically present artificial intelligence as subdiscipline of computing sciences and a body of knowledge for solving "difficult" challenges. This practice defines AI (indirectly) by those challenges that were considered "difficult but solvable" at the time of writing the textbook. The set of these challenges however has consistently grown since AI's infancy. For example, during the 1970s challenges such playing chess or the *Jeopardy!* game competently were considered too difficult. Most experts did not expect these challenges to be met anytime soon — with some arguing that AI could never win in these games or reason effectively in ambiguous domains such as natural language and image recognition. Within decades, Deep Blue beat the world's chess champion and IBM Watson won against two world champions in a televised *Jeopardy!* game. By the time of writing this, clever app developers can turn mobile devices to grandmaster-level chess players, and Watson, which can acquire knowledge from "reading" standard text such as encyclopaedias and articles, has learned to diagnose many medical conditions as well as most doctors. Image recognition programs have so improved over the years that the CAPTCHA task[8] have become increasingly difficult for humans to read, and simultaneously easy for computers to read, until CAPTCHAs have largely lost their claim to be able to "tell Computers and Humans Apart". Bots disguising themselves as available, lovelorn women, game characters, and friends in trouble are increasingly indistinguishable from humans. Thus within seven decades quite a few challenges that seemed possibly insurmountable have been downgraded to child's play in a similar manner. What *artificial intelligence* means in practice has changed accordingly from a limited set of capabilities into one that seeks to approximate humans'.

---

[7] The study also found one main difference, in that the risks addressed in the 1975 Asilomar conference appeared to be "overwhelmingly immediate", unlike human-level AI, which is not expected less than fifteen years in the future.

Since the meaning of 'artificial intelligence' has so much evolved, experts use the terms Human-Level Artificial Intelligence (HLAI) and Artificial General Intelligence (AGI) with reference to that point in the discipline's progress in which humans are challenged across a sufficiently wide range of capabilities. AI experts therefore commonly take artificial intelligence to be defined in comparison with human intelligence. This leads to several problems: The first problem is that psychologists predominantly reduce human intelligence to one quotient measured using psychometric (IQ) tests. But the hypothesis of a single, generic capability (called the *g factor*) does not explain for other forms of human intelligence such as creative genius, strategic talent, and social competence. It also does not explain non-human intelligence, although the abilities to use tools and plan ahead were demonstrated by many animal species. Obviously the *g factor* theory does not help us understand alien intelligence[9]. In conclusion, canonical accounts of human intelligence have not helped us measure artificial and nonhuman intelligence.

Legg and Hutter (Legg and Hutter 2007) define Universal Intelligence, a metric[10] for measuring the level of intelligence of any agent in a given environment — human, animal, or artificial. The metric uses *Kolmogorov complexity* to measures the difficulty of problems an agent solves well, a measure which has proven hard if not impossible to establish. And the metric's applicability is yet to be tested empirically. Importantly, however, the theory of Universal Intelligence defines the intelligence of a given agent *as a function of the environment in which it operates*. More precisely, the intelligence of an agent in a given environment is proportional to the agent's ability to maximize the utility of solving those problems which the environment is more likely to pose. Thus, an agent exhibits high intelligence in environments that happen to maximize the utility of those problems which the agent competently solves, and at once also exhibit low intelligence in other environments. This feature of Universal Intelligence sits well with our understanding of intelligence; an encouraging result. But Legg and Hutter's theory has so far not helped clarify what 'artificial intelligence' mean in more practical terms.

---

[8]  CAPTCHA stands for Completely Automated Public Turing test to tell Computers and Humans Apart, a test to determine whether an attempt for remote access is made by a human or by a bot.

[9]  That is, whatever it the Search for Extra Terrestrial Intelligence is searching for a sign of.

[10]  or rather a family of metrics

The difficulties to 'pin down' what AI means undermine our ability to assess its risks: we don't know which capabilities could (and could not) be developed in the near future, how long it will take to make available applications of each new capability, what is the likelihood of losing control over technology that will emerge from these applications, or under which circumstances if any the unintended consequences of future AI technology could pose an existential risk.

We explained the sense in which AI technologies can safely be said to have become more powerful, and that the trend may continue. But how far? Research in superintelligence (Bostrom 2014) have given us reasons to believe that the difference between apes and humans is dwarfed by gap between humans intelligence and superintelligence. Yet it is entirely unclear how much more intelligent will artificial agents become in the near future. Without further research we are left to guess whether no significant advances are likely or whether we should prepare ourselves to AI capabilities that are so advanced that it could 'turn the lights off' before we could blink. Such uncertainty is clearly undesirable.

## Conclusion (H) :
## The debate hasn't ended; it has barely begun

In 2012, after compiling the essays and rebuttals for the Singularity Hypotheses volume, we concluded that "the rapid growth in singularity research seems set to continue and perhaps accelerate" (Eden, Steinhart, et al. 2013). Since then prominent scientists have pushed to invest in this direction (Hawking et al. 2014; Future of Life Institute 2015), followed by sensational headlines and the controversy described above, which did not promote the debate but merely caricaturised and muddled it further. As the following questions remain open it is clear that the debate has barely began.

# OPEN QUESTIONS

Our investigation has also led us to conclude that the issues centre around a number of open questions, such as the following?

## Question 1:
## Can AI be controlled?

If loss of control over AI may indeed pose such a dangerous outcome then scientists feel compelled to ask how the risk can be eliminated or reduced. Research in this question has largely led to the conclusion that controlling the behaviour of agents that are more intelligent than their makers is difficult if not impossible (Bostrom 2014) — in particular if the agent learns to modify itself (Yampolskiy 2012).

One strategy of maintaining some control is commonly referred to as Friendly AI: research arising largely from the Machine Intelligence Research Institute (Yudkowsky 2001; Yudkowsky 2013) which focuses on methods of designing AIs whose behaviour would be considered 'friendly' or morally justified — for example by hardwiring artificial agents with our core values. A major difficulty with the approach is that it raises the notoriously tricky question of what exactly constitutes moral behaviour. For some people, morality implies equal distribution of wealth; to others, establishing an oppressive global caliphate. In other words, the notion of 'friendly' AI seems to require an objective theory of morality or a universal notion of 'good', a notoriously difficult goal if not a computationally intractable one (Brundage 2014). Furthermore, formal methods (Wing 1990; Hinchey et al. 2008) in software engineering have taught us that it is largely impossible to guarantee that a particular program implements a desired behaviour — assuming that behaviour (objective friendliness) can be established and formally defined.

Another strategy for controlling AIs seeks to confine artificial agents ("AI boxing") to a 'leak-proof' environment in a manner that could prevent undesired outcomes from materializing, for example, by sealing AIs in special hardware or by confining it to simulated environments. Vernor Vinge (Vinge 1993) has famously described the difficulty with the boxing approach using the example of superintelligent agent who can 'think' millions of times faster than its captives, and therefore could easily "come up with 'helpful advice' that would incidentally set [it] free".

A promising strategy for controlling AI suggests to restrict the use of artificial agents to answering Yes and No questions. The Oracle AI strategy (Bostrom 2014) minimizes the potential impact of AIs by limiting them to one capability, which is to answer questions either with Ttrue or False.

The question of controlling AI fits within the more general question of *mechanism design*, a body of work which in 2007 was awarded a Nobel Prize in economics for studying the optimal mechanisms to reach a certain goal. So far, this research indicates that such questions are difficult because information about individual preferences and available production technologies is distributed among many actors who may use their privately held information to further their own interests[11].

Finally, some suggested to outlaw research that may lead to artificial superintelligence (Joy 2000). However, even if regulators conclude that AI is so risky that its enormous benefits must be abandoned, banning it is unlikely to stop it. Rather, a ban is likely to push AI research underground, resulting in independent programmers and private labs conducting unmonitored experiments, which would be even less desirable.

A major difficulty with control strategies is that even if a foolproof method for controlling AIs is discovered it would have to be globally enforced, which is difficult because of the military incentives of nations and the monetary incentives of commercially motivated organization. In particular, totalitarian regimes and corporations may be tempted to loosen or remove AI controls for the purpose of gaining an advantage over their rivals, as demonstrated for example in the scenario of a global AI Arms Race (Istvan 2015; Future of Life Institute 2015).

## Question 2:
### AI or IA?

Analysing singularity hypotheses (Conclusion (B)) yielded the distinction between dystopian and utopian singularity scenarios, suggesting not a mutually exclusive circumstances but rather a race between artificial and posthuman superintelligence. The outcome of this race is therefore an open question whose answer may mark the difference between the best and the worst developments in human history. The question of who will 'win the race' has neatly been summarized as: "Will our successors be our descendants?" (Pearce 2013), or more pithily[12], "AI or IA?"

---

[11] Royal Swedish Academy of Sciences Press Release: "The Prize in Economic Sciences 2007".

## Question 3:
### Can we prove AIs are becoming more intelligent?

Critics of artificial superintelligence ideas often reject the common perception that AI has ever made genuine breakthroughs. It is argued that each new capability is merely an insignificant application of a pre-existing technology, and the systems which demonstrate them have remained narrow, rigid, and brittle (Allen 2011). One of the questions central to this debate is therefore whether one could quantify the intellectual advantage of newer AIs over systems built by older principles. A metric that can measure the intelligence of different AIs effectively might provide the means to find out how much more intelligent AI has become, and possibly improve our understanding of the way it is going.

# SUMMARY

Our central conclusion is that the controversy over the future of AI should be replaced by a reasoned and well-informed debate in the questions this technology raises, as the issues are much too important to —

— remain confused & misinformed
— be derailed by fantasy the likes of Hollywood's *Skynet*
— be left to academics & experts
— or be merely speculated about by pundits, amateurs, and opinionated philistines.

# ACKNOWLEDGEMENTS

The author wishes to thank Tony Willson for his encouragement and Mary J. Anna for her inspiration. This research was self-funded.

---

[12] George Dvorsky, "Humans With Amplified Intelligence Could Be More Powerful Than AI". *io9*, 22 May 2013.